%% file: rlc.tex
\documentclass[10pt]{article} 

\usepackage[accepted]{rlj} 

\usepackage{amssymb}            
\usepackage{graphicx}           
\usepackage{url}                

\usepackage{algorithmic}
\usepackage{mathtools}

\usepackage{physics}
\usepackage{mathrsfs}
\mathtoolsset{showonlyrefs}
\usepackage{etoolbox}
\usepackage{makecell}
\usepackage{booktabs}
\usepackage{lipsum}

\newcommand{\fS}{\mathcal{S}}
\newcommand{\fA}{\mathcal{A}}

\newcommand{\fY}{\mathcal{Y}}

\newcommand{\fT}{\mathcal{T}}

\newcommand{\fF}{\mathcal{F}}
\newcommand{\fO}{\mathcal{O}}

\newcommand{\R}{\mathbb{R}}
\newcommand{\indot}[2]{{\left<#1, #2\right>}}
\newcommand{\E}{\mathbb{E}}

\newcommand{\ns}{{\abs{\fS}}}
\newcommand{\na}{{\abs{\fA}}}

\newcommand{\explain}[1]{\tag*{(#1)}}

\newcommand{\tb}[1]{{\textbf{#1}}}

\usepackage{xcolor}
\definecolor{keywordcolor}{HTML}{4169e1}   %
\definecolor{tacticcolor}{HTML}{4169e1}    %
\definecolor{commentcolor}{HTML}{2e8b57}   %
\definecolor{symbolcolor}{HTML}{f75394}    %
\definecolor{sortcolor}{HTML}{4169e1}      %
\definecolor{attributecolor}{HTML}{f75394} %
\definecolor{lg}{gray}{0.95}

\newtheorem{theorem}{Theorem}[section]
\newtheorem{lemma}[theorem]{Lemma}

\newtheorem{assumption}[theorem]{Assumption}

\allowdisplaybreaks

\usepackage{comment}
\usepackage[final]{minted}
\setminted{
  frame=lines,
  framesep=1mm,
  fontsize=\scriptsize,
  baselinestretch=0.9,
  breaklines,
  autogobble,
  tabsize=2,
  vspace=-5pt,
}
\usepackage{MnSymbol}

\title{Towards Formalizing Reinforcement Learning \\Theory: A Robbins-Siegmund Approach}

\setrunningtitle{Towards Formalizing Reinforcement Learning Theory}

\author{Shangtong Zhang}

\emails{shangtong@virginia.edu}

\affiliations{
\textbf{Department of Computer Science, University of Virginia}
}

\contribution{
    The first formal verification of the almost sure convergence of $Q$-learning and linear TD with Markovian samples.
    }
    {
        Some results of dynamic programming and RL are formalized in existing works 
        \citep{vajjha2021certrl,vajjha2022formalization,chevallier2021formalising,schafeller2022formally,chevallier2024verification,schaffeler2025formally}. 
        But those works are far from formally verifying almost sure convergence of $Q$-learning and linear TD with Markovian samples.
        Some recent works \citep{li2024formalization,li2025formalizationa,li2025formalizationb,sonoda2025lean} formalize generalization bounds (via Rademacher complexity) and optimality of a few first-order optimization methods.
        They are also far from RL.
    }

\contribution{
    The formalization of a reusable Lean infrastructure based on the Robbins-Siegmund theorem for convergence of RL algorithms, instantiated here for linear TD and Q-learning to show effectiveness.
    }
    {
        Historically, the almost sure convergence analysis of RL algorithms is focused on ODE-based approaches \citep{benveniste1990MP,kushner2003stochastic,borkar2009stochastic,liu2025ode}.
        However, the current infrastructure in Mathlib does not support an easy formalization of ODE-based approaches.
        Instead, this work formalizes a more recent framework \citep{qian2024almost} based on the Robbins-Siegmund theorem.
        We demonstrate the effectiveness of this framework by using this framework to formalize the almost sure convergence of $Q$-learning and linear TD with Markovian samples.
        We argue that this framework is highly extensible because this framework has been used to provide asymptotic and nonasymptotic analyses of many RL algorithms in many different modes (e.g., almost surely, with high probability, in $\mathcal{L}^p$) \citep{qian2024almost,liu2025extensions}.
        We do note that those perceived extensions are not formalized yet and here we just argue for the extensibility.
    }

\keywords{RL Theory, Formal Verification, Lean} 

\summary{
    In this paper, we formalize the almost sure convergence of $Q$-learning and linear temporal difference (TD) learning with Markovian samples using the Lean 4 theorem prover based on the Mathlib library. $Q$-learning and linear TD are among the earliest and most influential reinforcement learning (RL) algorithms. The investigation of their convergence properties is not only a major research topic during the early development of the RL field but also receives significant attention nowadays. This paper formally verifies their almost sure convergence in a unified framework based on the Robbins-Siegmund theorem. The framework developed in this work can potentially be extended to convergence rates and other modes of convergence. This work thus makes an important step towards fully formalizing convergent RL results. 
}

\begin{document}

\makeCover  
\maketitle  

\begin{abstract}
\input{abstract.tex}
\end{abstract}

\input{body.tex}

\newpage


\bibliography{bibliography.bib}
\bibliographystyle{rlj}

\beginSupplementaryMaterials

\input{appendix.tex}


\end{document}

%% file: abstract.tex
In this paper, we formalize the almost sure convergence of $Q$-learning and linear temporal difference (TD) learning with Markovian samples using the Lean 4 theorem prover based on the Mathlib library. $Q$-learning and linear TD are among the earliest and most influential reinforcement learning (RL) algorithms. The investigation of their convergence properties is not only a major research topic during the early development of the RL field but also receives significant attention nowadays. This paper formally verifies their almost sure convergence in a unified framework based on the Robbins-Siegmund theorem. The framework developed in this work can potentially be extended to convergence rates and other modes of convergence. This work thus makes an important step towards fully formalizing convergent RL results. The code is available at \url{https://github.com/ShangtongZhang/rl-theory-in-lean}.

%% file: body.tex

\section{Introduction}
\label{sec intro}

Narrowly speaking, reinforcement learning (RL, \citet{sutton2018reinforcement}) is a framework for solving sequential decision making problems via trial and error. 
$Q$-learning \citep{watkins1989learning,watkins1992q} and linear temporal difference (TD) learning \citep{sutton1988learning} are among the earliest and most influential RL algorithms.
The investigation of their convergence property constitutes an important research topic in the RL theory community
\citep{watkins1989learning,watkins1992q,dayan1992convergence,jaakkola1993convergence,tsitsiklis1994asynchronous,tsitsiklis1997analysis,kearns1998finite,tsitsiklis1999average,even2003learning,azar2011speedy,beck2012error,shah2018q,bhandari2018finite,lakshminarayanan2018linear,srikant2019finite,lee2019unified,qu2020finite,li2020sample,chen2024lyapunov,li2024q,meyn2024projected,liu2025linearq,xie2025finite,liu2025extensions,blaser2026almost,blaser2026asymptotic,wang2024almost}.

We, however, argue that RL convergence proofs are usually delicate for two reasons.
\tb{First}, the almost sure convergence of RL algorithms is usually established through the ODE approach \citep{benveniste1990MP,kushner2003stochastic,borkar2009stochastic,borkar2025ode,liu2025ode}.
For example,
the seminal work \citet{tsitsiklis1997analysis} that establishes the almost sure convergence of linear TD relies on an ODE based stochastic approximation result in \citet{benveniste1990MP}.
The ODE based approach is full of details and bug-prone.
For example, \citet{degris2012off} investigate the convergence of off-policy actor critic algorithms through an ODE based approach, but as suggested by their erratum,
one major result in their peer-reviewed accepted version is entirely wrong. 
\citet{wan2020learning} investigate the convergence of average reward RL algorithms and point out that an earlier peer-reviewed accepted work gives a pseudoproof of the major result of \citet{wan2020learning}.
Even a well-established textbook makes gaps.
For example, the second version of \citet{borkar2009stochastic} states (and fixes) a major gap in its first version.
Those are publicly documented gaps (with or without fixes),
with more gaps hidden and only known to experts as folklore.
\tb{Second}, RL theory is typically formulated in the framework of Markov Decision Process (MDP, \citet{bellman1957markovian,puterman2014markov}).
To rigorously study the convergence of stochastic iterates inside the MDP framework,
one has to first construct the probability space for infinite length trajectories of the MDP (i.e., sample paths). 
This inevitably requires using the Ionescu-Tulcea theorem \citep{tulcea1949mesures}.
Consequently,
one has to verify the measurability and integrability of many functions in this probability space.
One also has to use a measure theoretic definition of conditional expectations with sub-$\sigma$-algebras in this probability space.
To reduce this definition to a more easy-to-use plain definition with conditional probability,
one again needs to verify the measurability and integrability of many related functions.
To our knowledge,
no prior RL theory work has gone through all those details to give a full rigor of their results.
Likely,
for finite state action MDPs,
those measurability and integrability can eventually be verified.
But for infinite ones,
there is a good chance that more assumptions are needed in existing results.
We regard formal verification as the ultimate approach for robustifying RL theory.
Accordingly,
this work develops the first formalization of the almost sure convergence of $Q$-learning and linear TD with Markovian samples on finite state action MDPs,
using the Lean 4 theorem prover \citep{moura2021lean} based on the Mathlib library \citep{mathlib2020}.\footnote{For readers unfamiliar with Lean and Mathlib, a brief introduction is available in Appendix~\ref{sec bkg lean}.}

Work has been done to formalize the basics of RL.
However, those are better categorized as formalizing dynamic programming instead of RL as there is no stochasticity in the algorithms they consider.
For example, \citet{vajjha2021certrl,chevallier2021formalising,schafeller2022formally,schaffeler2025formally} formalize the optimality of a few dynamic programming algorithms, e.g., (approximate) policy and value iteration, in Coq or Isabelle/HOL.
More related are \citet{vajjha2022formalization,chevallier2024verification},
which formalize Dvoretzky's theorem \citep{dvoretsky1955stochastic} for the almost sure convergence of a class of stochastic approximation algorithms in Coq and Isabelle/HOL, respectively.
Dvoretzky's theorem can be used to prove the almost sure convergence of some (arguably outdated) version of $Q$-learning (more details in Section~\ref{sec bkg}), but such a proof is never formalized in any prior work and Dvoretzky's theorem is incapable of analyzing linear TD.
By contrast, this paper is centered around modern techniques combining Lyapunov functions \citep{chen2024lyapunov} and the Robbins-Siegmund theorem \citep{robbins1971convergence} by using a skeleton iterates technique \citep{qian2024almost} to convert Markovian noise to Martingale difference noise,
which provides a unified framework for formalizing not only almost sure convergence but also high probability concentration, $\mathcal{L}^p$ convergence,
and the corresponding convergence rates.
Notable fellow projects within the machine learning community includes FoML \citep{sonoda2025lean} and Optlib \citep{li2024formalization,li2025formalizationa,li2025formalizationb},
both of which are in Lean.
FoML formalizes generalization bound by Rademacher complexity and is distant from the convergence of RL algorithms.
Optlib formalizes the optimality of a few first-order optimization methods (e.g., ADMM, gradient descent, Nesterov's accelerated methods).
While first-order optimization methods are closer to RL algorithms,
Optlib does not have any stochasticity either.
To summarize,
the focus on RL algorithms with Markov chain driven stochasticity distinguishes this paper from prior works of this kind.

The formalization can also serve as a high quality dataset for benchmarking LLM's reasoning and coding capability.
One example is  
\citet{yang2025formalml},
which develop a pipeline (centered around a new tactic in Lean) that can generate many subgoals from a complete Lean proof.
Those subgoal completion problems can then be used to benchmark LLMs.

\noindent \textbf{LLM statements.}
The formalization done in this paper greatly benefits from LLMs (specifically, Gemini and ChatGPT) in three ways.
First,
LLM serves as a personalized tutor that significantly bends the notoriously sharp learning curve of Lean.
Second,
LLM serves as a powerful search engine that can effectively retrieve corresponding lemmas from Mathlib based on natural language descriptions.
Third, LLM can complete some small lemmas automatically in one trial.
The auto-completion powered by Copilot is also very helpful in refactoring the implementation.
In this sense,
this project can be regarded as a collaboration between humans and AI
where humans dominate the collaboration.
With the help of LLMs,
we are able to complete this project in three months in part-time starting with zero knowledge of Lean.
A natural question for better gauging the contribution of this work is then:
  \emph{can LLM complete this project alone or with little help from humans?}
Our answer is negative as of Nov 2025.
First, from our own experience of interacting with LLMs during the project,
we frequently see hallucinations and significant incapacities of LLMs.
More importantly,
LLMs are mostly good at proving theorem statements that are already formalized and such proofs can be checked automatically.
However, we believe reliably and precisely converting theorem statements from natural language to formal statements and designing an extensible framework is still beyond the reach of current LLMs. 
Second,
the recent Gauss agent \citep{mathinc2025} achieves an important milestone in automated formalization by completely formalizing the strong prime number theorem.
In addition to the huge amount of computation the Gauss agent consumes,
it still receives significant input from humans in two ways.
First,
the Gauss agent does not start from scratch.
Instead, 
it starts from some important milestones (towards formalizing the strong prime number theorem) made by humans.
Second,
the Gauss agent relies on an 83-page human-written blueprint as a roadmap for formalization.
This blueprint is iterated multiple times by humans based on the progress and failures the agent makes.
Part of the blueprint is very detailed.
For example,
it can contain trivial lemmas such as the absolute value of a positive real is itself. 

\section{Background}
\label{sec bkg}

\textbf{Notations.} For $x, y \in \R^d$, we use $\indot{\cdot}{\cdot}$ to denote the inner product in Euclidean space, i.e., $\indot{x}{y} = x^\top y$. 
We use $\norm{x}_p \doteq \qty(\sum_i \abs{x_i}^p)^{1/p}$ to denote the $\ell_p$ norm and $\norm{x}_\infty \doteq \max_i \abs{x_i}$ to denote the infinity norm.
We overload the vector norms to also denote the induced matrix norms.

We consider an infinite horizon MDP with a finite state space $\fS$, a finite action space $\fA$, a reward function $r : \fS \times \fA \to \R$, a transition function $p : \fS \times \fS \times \fA \to [0, 1]$, an initial distribution $p_0 : \fS \to [0, 1]$,
and a discount factor $\gamma \in [0, 1)$.
At time step $0$,
an initial state $S_0$ is sampled from $p_0$.
At time $t$ and state $S_t$,
an action $A_t$ is sampled from $\pi(\cdot | S_t)$,
where $\pi : \fA \times \fS \to [0, 1]$ is the policy. 
A successor state $S_{t+1}$ is then sampled from $p(\cdot | S_t, A_t)$ and a reward $R_{t+1} \doteq r(S_t, A_t)$ is generated.
The state value function is defined as $v_\pi(s) \doteq \E\qty[\sum_{i=0}^\infty \gamma^i R_{t+i+1} | S_t = s]$
and the action value function is defined as
$q_\pi(s, a) \doteq \E\qty[\sum_{i=0}^\infty \gamma^i R_{t+i+1} | S_t = s, A_t = a]$.
Estimating $v_\pi$ is one fundamental task in RL,
called policy evaluation.
Another fundamental task is control,
the goal of which is to find an optimal policy $\pi_*$  
such that $q_{\pi_*}(s, a) \geq q_\pi(s, a)\, \forall \pi, s, a$.
There can be multiple optimal policies but all of them must share the same action value function,
denoted as $q_*$ and called the optimal action value function,
which is the unique fixed point of the Bellman optimality operator $\fT_* : \R^{\ns\times \na} \to \R^{\ns \times \na}$
 defined as $(\fT_* q)(s, a) \doteq r(s, a) + \gamma \sum_{s'} p(s' |s, a) \max_{a'} q(s', a')$.

TD is one of the most well-received algorithms for policy evaluation,
which 
estimates $v_\pi$ via stochastic iterates $\qty{v_t \in \R^\ns}$ generated as
$v_{t+1}(s) = v_t(s) + \alpha_t (R_{t+1} + \gamma v_t(S_{t+1}) - v_t(S_t))\mathbb{I}_{s = S_t}$,
where $\qty{\alpha_t}$ is a sequence of deterministic step sizes and $\mathbb{I}$ is the indicator function.
$Q$-learning is one of the most well-received algorithms for control,
which estimates $q_*$ via stochastic iterates $\qty{q_t \in \R^{\ns\times \na}}$ generated as
\begin{align}
    \label{eq q learning}
   \!\!\!\!\!\!\!\!\!\!\!\!\!\!\!\!\!\!\!\!\!\!\!\!\!\!\!\!q_{t+1}(s, a) = q_t(s, a) + \alpha_t (R_{t+1} + \gamma \max_b q_t(S_{t+1}, b) - q_t(S_t, A_t)) \mathbb{I}_{(s, a) = (S_t, A_t)}.
   \tag{$Q$-learning}
\end{align}
Here we consider the case where the action $A_t$ is sampled from a fixed policy $\pi$.
It is well-known (e.g., \citet{qian2024almost}) that almost surely, $\lim_{t\to\infty} v_t = v_\pi$ and $\lim_{t\to\infty} q_t = q_*$.
Instead of using a look-up table $\qty{v_t}$ to store estimates of $v_\pi$, 
parameterized functions can also be used.
Particularly,
\citet{sutton1988learning} considers a linear parameterization.
Let $x : \fS \to \R^K$ be a feature function that maps a state $s$ to a $K$-dimensional feature.
We then use $x(s)^\top w$ to approximate $v_\pi(s)$,
where $w \in \R^K$ is a learnable weight.
Linear TD then generates iterates $\qty{w_t \in \R^K}$ as
\begin{align}
    \label{eq linear td}
    w_{t+1} = w_t + \alpha_t (R_{t+1} + \gamma x(S_{t+1})^\top w_t - x(S_t)^\top w_t) x(S_t).
    \tag{Linear TD}
\end{align}
It is well-known (e.g., \citet{tsitsiklis1997analysis}) that $\lim_{t\to\infty} w_t = w_*$ a.s., where $w_*$ is the TD fixed point.
To define $w_*$,
we use $X \in \R^{\ns \times K}$ to denote the feature matrix whose $s$-th row is $x(s)^\top$,
use $P_\pi \in \R^{\ns \times \ns}$ to denote the transition matrix of the Markov chain induced by $\pi$ such that $P_\pi(s, s') = \sum_a \pi(a|s) p(s'|s, a)$,
use $r_\pi \in \R^{\ns}$ to denote the reward vector such that $r_\pi(s) = \sum_a \pi(a|s) r(s, a)$,
and use $D_\pi \in \R^{\ns \times \ns}$ to denote the diagonal matrix whose diagonal term is the stationary state distribution $d_\pi \in \R^\ns$ of the Markov chain $\qty{S_t}$ induced by $\pi$.
Then we have $w_* = -A^{-1}b$,
where $A\doteq X^\top D_\pi(\gamma P_\pi - I)X$ and $b \doteq X^\top D_\pi r_\pi$.

The goal of this paper is thus to formally prove that $\lim_{t\to\infty} q_t = q_*$ a.s. and $\lim_{t\to\infty} w_t = w_*$ a.s.
We note that for~\eqref{eq linear td}, 
we follow \citet{tsitsiklis1997analysis} and consider a Markov Reward Process (MRP) setup,
i.e.,
we use $R_{t+1} \doteq r_\pi(S_t)$ directly in~\eqref{eq linear td}.
We also note that 
earlier works of $Q$-learning use a different update rule that replaces $\alpha_t$ in~\eqref{eq q learning} with $\alpha_{\nu(S_t, A_t, t)}$,
where $\nu(s, a, t) \doteq \sum_{\tau = 0}^t \mathbb{I}_{(s, a) = (S_i, A_i)}$ is
a counter that counts the visit of $(s, a)$ up to time $t$.
This counter is rarely used by practitioners or the modern formulation of $Q$-learning \citep{sutton2018reinforcement} and \citet{antrobius2026divergence} show that such a counter is mostly a technical convenience to significantly simplify the proof.
We, therefore, do not consider this counter in our formalization.
The form~\eqref{eq q learning} is also what \citet{chevallier2024verification} proposed to formalize (but did not) after they formalized Dvoretzky's theorem.
\citet{chevallier2024verification} states that the formal proof of~\eqref{eq q learning} is very close after the formal proof of Dvoretzky's theorem.
We, however, argue that there is a gap.
If the counter-based step size was used in~\eqref{eq q learning},
then the convergence would follow easily from Dvoretzky's theorem.
But for the exact form of~\eqref{eq q learning},
one needs to additionally prove that $\forall (s, a), \sum_{t=0}^\infty \alpha_t \mathbb{I}_{(s, a) = (S_t, A_t)} = \infty$ a.s. to use Dvoretzky's theorem.
This is true under moderate assumptions on the Markov chain but highly nontrivial to formalize, especially given that \citet{chevallier2024verification} does not have a measure theoretic formalization of the probability space of sample paths of the Markov chain.

\section{Formal Theorem Statements}
\label{sec statement}
We now describe our formalization of the theorem statements. We start with the almost sure convergence of~\eqref{eq linear td}. To this end,
we first define stochastic vectors on a finite state space $\fS$ and the corresponding row stochastic matrix.
\begin{minted}[escapeinside=@@]{lean}
variable {S : Type u} [Fintype S]
class StochasticVec (x : S @$\to$@ @$\R$@) where
  nonneg : @$\forall$@ s, 0 @$\le$@ x s
  rowsum : @$\sum$@ s, x s = 1
class RowStochastic (P : Matrix S S @$\R$@) where
  stochastic: @$\forall$@ s, StochasticVec (P s)
\end{minted}
We then define irreducibility and aperiodicity of row stochastic matrices.
\begin{minted}[escapeinside=@@]{lean}
class Irreducible (P : Matrix S S @$\R$@) [RowStochastic P] where
  irreducible : @$\forall$@ i j, @$\exists$@ n : @$\mathbb{N}$@, 0 < (P ^ n) i j
class Aperiodic (P : Matrix S S @$\R$@) [RowStochastic P] where
  aperiodic : @$\forall$@ i, FiniteGCDOne (return_times P i)
\end{minted}
Irreducibility and aperiodicity together imply Doeblin minorization after sufficient powers.
\begin{minted}[escapeinside=@@]{lean}
class DoeblinMinorization (P : Matrix S S @$\R$@) [RowStochastic P] where
  minorize : @$\exists$@ (@$\epsilon$@ : @$\R$@) (@$\nu$@ : S @$\to$@ @$\R$@), 0 < @$\epsilon$@ @$\land$@ @$\epsilon$@ < 1 @$\land$@ StochasticVec @$\nu$@ @$\land$@ @$\forall$@ i j, P i j @$\ge$@ @$\epsilon$@ * @$\nu$@ j
theorem smat_minorizable_with_large_pow [Nonempty S] (P : Matrix S S @$\R$@)
  [RowStochastic P] [Irreducible P] [Aperiodic P] : @$\exists$@ N, 1 @$\le$@ N @$\land$@ DoeblinMinorization (P ^ N)
\end{minted}
The corresponding matrix power operator is then a contraction in the simplex.
\begin{minted}[escapeinside=@@]{lean}
theorem smat_contraction_in_simplex (P : Matrix S S @$\R$@) [RowStochastic P] 
  [DoeblinMinorization P] : @$\exists$@ K, 0 < K @$\land$@ ContractingWith K (smat_as_operator P)    
\end{minted}
This allows us to invoke Banach's fixed point theorem to conclude the existence and uniqueness of the stationary distribution
as well as the geometric mixing property.
\begin{minted}[escapeinside=@@]{lean}
theorem stationary_distribution_uniquely_exists (P : Matrix S S @$\R$@) [RowStochastic P] 
  [Aperiodic P] [Irreducible P] : @$\exists$@! @$\mu$@ : S @$\to$@ @$\R$@, StochasticVec @$\mu$@ @$\land$@ Stationary @$\mu$@ P    
instance (P : Matrix S S @$\R$@) [RowStochastic P] [Aperiodic P] [Irreducible P]
  : GeometricMixing P
\end{minted}
Having defined the stationary distribution,
we are finally ready to formalize the TD fixed point $w_*$.
\begin{minted}[escapeinside=@@]{lean}
abbrev E (d : @$\mathbb{N}$@) := EuclideanSpace @$\R$@ (Fin d)
noncomputable def LinearTDSpec.td_fixed_point : E d := - spec.A@$^{-1}$@ *@$_v$@ spec.b
\end{minted}
We now describe how we construct the sample path probability space.
To this end, we first define a time-homogeneous Markov chain using probability kernels from Mathlib.
\begin{minted}[escapeinside=@@]{lean}
structure HomMarkovChainSpec (S : Type u) [MeasurableSpace S] where
  kernel : Kernel S S
  markov_kernel : IsMarkovKernel kernel
  init : ProbabilityMeasure S
\end{minted}
Notably, in Mathlib, \texttt{IsMarkovKernel} only means the kernel is a probability kernel and has nothing to do with the Markovian property of a Markov chain.
We then generate the probability measure on the sample path space $S^\infty$.
\begin{minted}[escapeinside=@@]{lean}
noncomputable def traj_prob (M : HomMarkovChainSpec S) : ProbabilityMeasure (@$\mathbb{N}$@ @$\to$@ S) 
\end{minted}
This is done by invoking the Ionescu-Tulcea theorem in Mathlib \citep{marion2025formalization},
which uses a constructive way to prove the existence of a probability measure on $S^\infty$ that coincides with the iterated application of the transition kernel $P$ on any partial sample path with finite length. 
The best practice (e.g., \citet{tsitsiklis1997analysis}) for analyzing the convergence of linear TD in existing literature is to consider the augmented Markov $Y_t \doteq (S_t, S_{t+1})$.
We follow this and eventually realize \texttt{HomMarkovChainSpec} with a state space $\fY \doteq \fS \times \fS$.
The corresponding transition kernel, described here in a matrix form for simplicity, is then $P_\fY((s_0, s'_0), (s_1, s'_1)) = \mathbb{I}_{s_1 = s_0'} P_\pi (s_1, s_1')$.
The Ionescu-Tulcea theorem then generates a probability measure on $(S \times S)^\infty$.
Given a sample path $\omega \in (S \times S)^\infty$,
the algorithm~\eqref{eq linear td} is then defined as
\begin{minted}[escapeinside=@@]{lean}
noncomputable def LinearTDSpec.update (w : E d) (y : S @$\times$@ S) : E d :=
  (spec.r y.1 + spec.@$\gamma$@ * @$\llangle$@spec.x y.2, w@$\rrangle$@ - @$\llangle$@spec.x y.1, w@$\rrangle$@) @$\cdot$@ spec.x y.1
variable {w : @$\mathbb{N}$@ @$\to$@ (@$\mathbb{N}$@ @$\to$@ (S @$\times$@ S)) @$\to$@ E d}
class LinearTDIterates where
  init : @$\forall$@ @$\omega$@, w 0 @$\omega$@ = spec.w@$_0$@
  step : @$\forall$@ n @$\omega$@, w (n + 1) @$\omega$@ = w n @$\omega$@ + spec.@$\alpha$@ n @$\cdot$@ spec.update (w n @$\omega$@) (@$\omega$@ (n + 1))
\end{minted}
We now formalize the almost sure convergence of linear TD as
\begin{minted}[escapeinside=@@]{lean}
theorem ae_tendsto_of_linearTD_markov
  {@$\nu$@ : @$\R$@} (h@$\nu$@ : @$\nu$@ @$\in$@ Set.Ioo (2 / 3) 1) (hw : LinearTDIterates (spec := spec) (w := w))
  (h@$\alpha$@ : spec.@$\alpha$@ = fun n : @$\mathbb{N}$@ => inv_poly @$\nu$@ 2 n) :
  @$\forall$@@$^m$@ @$\omega$@ @$\partial$@ spec.markov_samples, Tendsto (fun n => w n @$\omega$@) atTop (@$\mathcal{N}$@ spec.td_fixed_point) 
\end{minted}
Precisely, what we formalize is
\begin{theorem}
    \label{thm linear td}
Let the finite Markov chain $\qty{S_t}$ be irreducible and aperiodic. 
Let $X$ have a full column rank.
Let the step size be $\alpha_t = \frac{1}{(t + 2)^\nu}$ with $\nu \in (2/ 3 , 1)$.
Then the iterates $\qty{w_t}$ generated by~\eqref{eq linear td} with $R_{t+1} \doteq r_\pi(S_t)$ satisfy that $\lim_{t\to\infty} w_t = w_*$ a.s.
\end{theorem}
The particular choice of $\nu \in (2/3, 1)$ is an artifact of our proof technique and we shall revisit this in the next section.
We similarly formalize the almost sure convergence of $Q$-learning as
\begin{minted}[escapeinside=@@]{lean}
theorem ae_tendsto_of_QLearning_markov
  {@$\nu$@ : @$\R$@} (h@$\nu$@ : @$\nu$@ @$\in$@ Set.Ioo (2 / 3) 1) (hq : QLearningIterates (spec := spec) (q := q))
  (h@$\alpha$@ : spec.@$\alpha$@ = fun n : @$\mathbb{N}$@ => inv_poly @$\nu$@ 2 n) :
  @$\forall$@@$^m$@ @$\omega$@ @$\partial$@ spec.MRP.markov_samples, Tendsto (fun n => q n @$\omega$@) atTop (@$\mathcal{N}$@ spec.optimal_q)
\end{minted}
Precisely, what we formalize is
\begin{theorem}
    \label{thm q}
For any fixed policy $\pi$,
let the induced finite Markov chain $\qty{(S_t, A_t)}$ be irreducible and aperiodic. 
Let the step size be $\alpha_t = \frac{1}{(t + 2)^\nu}$ with $\nu \in (2/ 3 , 1)$.
Then the iterates $\qty{q_t}$ generated by~\eqref{eq q learning} satisfy that $\lim_{t\to\infty} q_t = q_*$ a.s.
\end{theorem}
We also formalize the two almost sure convergences under i.i.d. samples,
where more step sizes are allowed as long as the step sizes satisfy the Robbins-Monro condition\footnote{In this work, we say a sequence $\qty{\alpha_t}$ satisfies the Robbins-Monro condition if $0 < \alpha_t, \sum_t \alpha_t = \infty, \sum_t \alpha_t^2 < \infty$.}. 

\section{Formal Theorem Proofs}
\label{sec proof}
The canonical almost sure convergence analysis of linear TD relies on ODE-based methods \citep{benveniste1990MP,kushner2003stochastic,borkar2009stochastic,borkar2025ode,liu2025ode}.
However,
our evaluation is that those ODE-based approaches are not ready for formalization yet. 
The main reason is that Mathlib has only very few results about ODE and control theory.
Instead, this paper uses a more modern approach based on the Robbins-Siegmund theorem.
We now elaborate more on our roadmap,
which is largely based on \citet{chen2024lyapunov,qian2024almost}.
Both~\eqref{eq q learning} and~\eqref{eq linear td} can be rewritten in the form of
\begin{align}
    \label{eq raw update}
  w_{t+1} = w_t + \alpha_t (F(w_t, Y_{t+1}) - w_t).
\end{align}
For~\eqref{eq linear td},
we have $Y_{t+1} \doteq (S_t, S_{t+1})$ and 
$F(w, (s, s')) = (r_\pi(s) + \gamma x(s')^\top w - x(s)^\top w) x(s) + w$.
For~\eqref{eq q learning},
we have $Y_{t+1} \doteq (S_t, A_t, S_{t+1}, A_{t+1})$ and
$F(q, (s, a, s', a'))(s_0, a_0) = (r(s, a) + \gamma \max_{b} q(s', b) - q(s, a))\mathbb{I}_{(s, a) = (s_0, a_0)} + q(s_0, a_0)$.
Notably, $F$ actually does not depend on the argument $a'$.
It is included here only for a unified proof implementation for both~\eqref{eq q learning} and~\eqref{eq linear td}.
Let $f(w)$ be the expectation of $F(w, \cdot)$ w.r.t. the stationary distribution.
We then have 
\begin{align}
  \label{eq unified update iid}
  w_{t+1} = w_t + \alpha_t (f(w_t) - w_t) + \alpha_t (F(w_t, Y_{t+1}) - f(w_t)).
\end{align}
This motivates us to study the iterates described below.
\begin{minted}[escapeinside=@@]{lean}
class Iterates (x : @$\mathbb{N}$@ @$\to$@ @$\omega$@ @$\to$@ E d)
  (x@$_0$@ : E d) (f : E d @$\to$@ E d) (e@$_1$@ e@$_2$@ : @$\mathbb{N}$@ @$\to$@ @$\omega$@ @$\to$@ E d) (@$\alpha$@ : @$\mathbb{N}$@ @$\to$@ @$\R$@) where
  init : @$\forall$@ @$\omega$@, x 0 @$\omega$@ = x@$_0$@
  step : @$\forall$@ n @$\omega$@, x (n + 1) @$\omega$@ =
    x n @$\omega$@ + (@$\alpha$@ n) @$\cdot$@ (f (x n @$\omega$@) - x n @$\omega$@) + e@$_1$@ (n + 1) @$\omega$@ + e@$_2$@ (n + 1) @$\omega$@  
\end{minted}
Precisely, let $\Omega$ be a set equipped with a $\sigma$-algebra, let $\mu$ be a probability measure, and let \\ $\qty{x_n, e_{1, n}, e_{2, n} : \Omega \to \R^d}$ be a sequence of measurable functions satisfying $\forall \omega \in \Omega$
\begin{align}
  \label{eq iterates 1}
  x_{n+1}(\omega) = x_n(\omega) + \alpha_n (f(x_n(\omega)) - x_n(\omega)) + e_{1, n+1}(\omega) + e_{2, n+1}(\omega).
\end{align}
We then study the convergence of $x_n$ under assumptions on the noise terms $e_{1,n}$ and $e_{2, n}$.
Here $f : \R^d \to \R^d$ is the function of interest that has a fixed point $x_*$ such that $f(x_*) = x_*$.
Our goal is thus to show $\lim_{n\to\infty} x_n(\omega) = x_*$ for almost all $\omega$.
We make some basic assumptions on $f$ and $\alpha_n$.
\begin{assumption}
  \label{ass lr}
  $\qty{\alpha_n}$ satisfy the Robbins-Monro condition and $f$ is Lipschitz continuous.
\end{assumption}
We further assume the existence of a Lyapunov function $\phi : \R^d \to [0, \infty)$ that satisfies
\begin{assumption} 
  \label{ass phi}
  $\forall x, y$
  \begin{itemize}
    \item[(a)] $\phi(y) \leq \phi(x) + \indot{\nabla \phi(x)}{y - x} + C \norm{y - x}_2^2$
    \item[(b)] $\phi(x) \geq 0$ and $\phi(x) = 0 \iff x = 0$
    \item[(c)] $\indot{\nabla \phi(x)}{x} = C \phi(x), \sum_i \abs{(\nabla \phi(x))_i} \abs{y_i} \leq C \sqrt{\phi(x)}\sqrt{\phi(y)}, \norm{x}_2 \leq C \sqrt{\phi(x)}, \sqrt{\phi(x)} \leq C \norm{x}_2$
    \item[(d)] $\indot{\nabla \phi(x - x_*)}{f(x) - x} \leq -\eta \phi(x - x_*) $
  \end{itemize}
\end{assumption}
Here $C$ just denotes the existence of some nonnegative constants and $C$ does not need to be the same for each of its appearances.
Notably, $\eta$ needs to be strictly positive.
Assumption~\ref{ass phi} (a) essentially says that $\phi$ is smooth.
Assumptions~\ref{ass phi} (b) \& (c) says that $\phi$ needs to behave like a squared norm.
Assumption~\ref{ass phi} (d) says the update direction $f(x) - x$ should decay the Lyapunov function.
We prove that $\phi(x) = \frac{1}{2}\norm{x}_p^2$ satisfies (a), (b), and (c) for $p \geq 2$.
When $p=2$,
(d) is verified for the $f$ corresponding to~\eqref{eq linear td}.
For linear TD, it can be computed that
$f(w) = Aw + b + w$.
Then 
\begin{align}
  \textstyle \indot{\nabla \frac{1}{2}\norm{w - w_*}^2_2}{f(w) - w} = \indot{w - w_*}{Aw + b} = \indot{w - w_*}{A(w - w_*)} \leq -\eta \norm{w - w_*}^2_2,
\end{align}
where the second equality is due to $Aw_* + b = 0$ and the last inequality is due to that $A$ is negative definite.
When $p$ is sufficiently large,
(d) is verified for the $f$ corresponding to~\eqref{eq q learning}.
Specifically, for~\eqref{eq q learning}, 
define a weighted Bellman optimality operator as $(\fT_*' q)(s, a) \doteq d_{\pi_q}(s){\pi_q}(a|s)\qty[(\fT_* q)(s, a) - q(s, a)] + q(s, a)$.
For this weighted Bellman optimality operator,
the behavior policy $\pi$ is allowed to depend on the action value estimation $q$. 
\citet{liu2025linearq} prove that $\fT_*'$ is a pseudo-contraction,
i.e.,
there exists a $\gamma' \in [0, 1)$ such that $\forall q, \norm{\fT_*' q - q_*}_\infty \leq \gamma' \norm{q - q_*}_\infty$.
In this paper,
we consider the setup where the behavior policy is fixed so $\pi_q$ degenerates to $\pi$ directly.
For our $f$ corresponding to~\eqref{eq q learning},
it can be computed that
$f(q) = \fT_*' q$.
We then have (see Section~\ref{sec proof of eq q lyapunovCandidate} in the supplementary materials for a brief proof)
\begin{align}
  \label{eq q lyapunovCandidate}
  \textstyle \indot{\nabla \frac{1}{2} \norm{q - q_*}_p^2}{f(q) - q} 
  \leq& -(1 - \gamma'(\ns\na)^{1/p}) \norm{q - q_*}^2_p. 
\end{align}
For sufficiently large $p$,
we have $\eta \doteq (1 - \gamma'(\ns\na)^{1/p}) > 0$.
Back to~\eqref{eq iterates 1}, we now make assumptions on the growth of the noise terms.
\begin{assumption} 
  \label{ass e growth}
  There exists $C \geq 0$ such that $\forall n$ and almost every $\omega$,
  \begin{align}
    \norm{e_{1, n+1}(\omega)}_2 \leq C \alpha_n (1 + \norm{x_n(\omega)}^2), \,
    \norm{e_{2, n+1}(\omega)}_2 \leq C \alpha_n^2 (1 + \norm{x_n(\omega)}^2).
  \end{align}
\end{assumption}
We are now able to prove a recursive error bound for almost every $\omega$.
\begin{lemma}
  \label{thm fundamental ineq}
  Let Assumptions~\ref{ass lr} -~\ref{ass e growth} hold.
  Then there exist some constants $C_1 > 0, C_2 \geq 0$, and $n_0 \geq 0$ such that $\forall n \geq n_0$ and for almost every $\omega$
  \begin{align}
      \label{eq fundamental ineq}
    \phi(x_{n+1}(\omega) - x_*) \leq (1 - C_1 \alpha_n) \phi(x_n(\omega) - x_*) + \indot{\nabla \phi(x_n(\omega) - x_*)}{e_{1, n+1}(\omega)}  + C_2 \alpha_n^2.
  \end{align}
\end{lemma}
We now further assume that $\qty{e_{1, n}}$ is a Martingale difference sequence.
\begin{assumption}
  \label{ass mds}
  $\qty{x_n}$ is adapted to a filtration $\qty{\fF_n}$ and $\E[e_{1, n+1} | \mathcal{F}_n] = 0$ a.s.
\end{assumption}
Here we recall that the conditional expectation $\E[e_{1, n+1} | \mathcal{F}_n]$ is the unique (up to nullset of $\mu$) function $\Omega \to \R^d$ such that for any $B \in \fF_n, \int_B \E\qty[e_{1, n+1} | \fF_n] \text{d}\mu = \int_B e_{1, n+1} \text{d}\mu$.
Taking conditional expectations on both sides of~\eqref{eq fundamental ineq} then generates that 
\begin{align}
  \E\qty[\phi(x_{n+1}(\omega) - x_*)|\fF_n] \leq (1 - C_1 \alpha_n) \E\qty[\phi(x_n(\omega) - x_*)|\fF_n] + C_2 \alpha_n^2 \qq{a.s.}
\end{align}
This means that the sequence of functions $\qty{\omega \mapsto \phi(x_n(\omega) - x_*)}$ is almost a supermartingale.
By a special case of the Robbins-Siegmund theorem formalized below,
\begin{minted}[escapeinside=@@]{lean}
theorem ae_tendsto_zero_of_almost_supermartingale
  (hAdapt : Adapted @$\mathcal{F}$@ f) (hfm : @$\forall$@ n, Measurable (f n)) (hfInt : @$\forall$@ n, Integrable (f n) @$\mu$@)
  (hfnonneg : @$\forall$@ n, 0 @$\le$@@$^m$@[@$\mu$@] f n) {T : @$\mathbb{N}$@ @$\to$@ @$\R$@} (hTpos   : @$\forall$@ n, 0 < T n)
  {hTsum : Tendsto (fun n => @$\sum$@ k @$\in$@ range n, T k) atTop atTop}
  {hTsqsum : Summable (fun n => (T n) ^ 2)}
  (hAlmostSupermartingale : @$\exists$@ C @$\ge$@ 0, @$\forall$@ n, 
    @$\mu$@[f (n + 1) | @$\mathcal{F}$@ n] @$\le$@@$^m$@[@$\mu$@] (fun @$\omega$@ => (1 - T n) * f n @$\omega$@ + C * T n ^ 2)) :
  @$\forall$@@$^m$@ @$\omega$@ @$\partial$@@$\mu$@, Tendsto (fun n => f n @$\omega$@) atTop (@$\mathcal{N}$@ 0)
\end{minted}
we obtain that $\lim_{n\to\infty} \phi(x_n(\omega) - x_*) = 0$ a.s.
Precisely, the version of the Robbins-Siegmund theorem and the stochastic approximation result we formalize so far are
\begin{theorem}
  [A special case of \citet{robbins1971convergence}]
  Let $\qty{z_n : \Omega \to \R}$ be a sequence of functions such that $z_n \geq 0$ a.s. and $z_n$ is integrable. Let $\qty{\fF_n}$ be a filtration such that $z_n$ is measurable by $\fF_n$.
  Let $\qty{T_n}$ be a sequence of deterministic reals satisfying the Robbins-Monro condition.
  If $\qty{z_n}$ is almost a supermartingale given $\qty{T_n}$ and some nonnegative constant $C$ in the sense that
    $\E\qty[z_{n+1}|\fF_n] \leq (1 - T_n) z_n + CT_n^2$ a.s.,
  then $\lim_{n\to\infty} z_n = 0$ a.s.
\end{theorem}
\begin{theorem}
  \label{thm mds}
  Let Assumptions~\ref{ass lr} -~\ref{ass mds} hold. Then $\qty{x_n(\omega)}$ in~\eqref{eq iterates 1} satisfy $\lim_{n\to\infty} x_n(\omega) = x_*$ a.s.
\end{theorem}
Looking back at~\eqref{eq iterates 1}, the roles of the two noise terms are clearer now.
The noise $e_1$ is larger (of $\fO(\alpha_n)$) but needs to be a Martingale difference sequence.
The noise $e_2$ is smaller (of $\fO(\alpha_n^2)$) but does not need to have other special properties.
We will shortly see how Markovian samples can fit into the two noise terms but
Theorem~\ref{thm mds} is already enough for the almost sure convergence of~\eqref{eq linear td} and~\eqref{eq q learning} with i.i.d. samples.
Specifically,
if $\qty{Y_t}$ is i.i.d. in~\eqref{eq unified update iid},
we can identify $e_{1, n+1}$ as $\alpha_t (F(w_t, Y_{t+1}) - f(w_t))$ and $e_{2, n+1}$ as 0 and then invoke Theorem~\ref{thm mds}.
To work with Markovian $\qty{Y_t}$,
we follow the skeleton iterates technique in \citet{qian2024almost},
which is essentially an improved version of a proof technique used in the proof of Proposition 4.8 of \citet{bertsekas1996neuro}.
We use $G(w, Y) \doteq F(w, Y) - w$ and $g(w) \doteq f(w) - w$ as shorthands.
We then consider a deterministic and strictly increasing sequence $\qty{t_m}_{m=0,1,\dots}$, called the anchors, with $t_0 = 0$ and $\lim_{m\to\infty}t_m = \infty$.
Telescoping~\eqref{eq raw update} then yields that for any $m$,
\begin{align}
    w_{t_{m+1}} =& \textstyle w_{t_m} + \sum_{t = t_m}^{t_{m+1}-1} \alpha_t G(w_t, Y_{t+1}) \\
    =&\textstyle w_{t_m} + \beta_m g(w_{t_m}) + \sum_{t=t_m}^{t_{m+1} - 1} \alpha_t \qty(G(w_{t_m}, Y_{t+1}) - \E\qty[G(w_{t_m}, Y_{t+1})| \fF_{t_m}]) \\
    &\textstyle+\sum_{t=t_m}^{t_{m+1} - 1} \alpha_t \qty(\E\qty[G(w_{t_m}, Y_{t+1}) | \fF_{t_m}] - g(w_{t_m}) + G(w_t, Y_{t+1}) - G(w_{t_m}, Y_{t+1})),
\end{align}
where $\beta_m \doteq \sum_{t=t_m}^{t_{m+1} - 1}\alpha_t$.
We can then in~\eqref{eq iterates 1} identify $x_n(\omega)$ as $w_{t_m}$, $\alpha_n$ as $\beta_m$, $\fF_n$ as $\fF_{t_m}$,
$e_{1, n+1}(\omega)$ as $\sum_{t=t_m}^{t_{m+1} - 1} \alpha_t \qty(G(w_{t_m}, Y_{t+1}) - \E\qty[G(w_{t_m}, Y_{t+1})| \fF_{t_m}])$,
and $e_{2,n+1}(\omega)$ as $\sum_{t=t_m}^{t_{m+1} - 1} \alpha_t \qty(\E\qty[G(w_{t_m}, Y_{t+1}) | \fF_{t_m}] - g(w_{t_m}) + G(w_t, Y_{t+1}) - G(w_{t_m}, Y_{t+1}))$.
Verifying Assumptions~\ref{ass lr} -~\ref{ass mds} and invoking Theorem~\ref{thm mds} then concludes that $\lim_{m\to\infty}w_{t_m} = w_*$ a.s.
A few Gronwall's inequalities further give $\lim_{t\to\infty}w_t = w_*$.
See \citet{qian2024almost} for a detailed proof in natural language.
Notably,
the anchors $\qty{t_m}$ need an additional property that $\alpha_{t_m} \leq C \beta_m^2$.
For this to hold,
we need $\nu \in (2/3, 1)$ in Theorems~\ref{thm linear td} \&~\ref{thm q}. 
For $\nu = 1$,
\citet{qian2024almost} have already also proved it. 
So it will be straightforward to formalize.
For $\nu \in (1/2, 2/3]$,
we need to either extend the results of \citet{qian2024almost} or resort to the canonical ODE based approach \citep{benveniste1990MP,kushner2003stochastic,borkar2009stochastic,borkar2025ode,liu2025ode},
among which our evaluation is that \citet{liu2025ode} is the most plausible to formalize and is perhaps the most powerful in terms of almost sure convergence.
For $\nu \in (0, 1/2]$,
the $\qty{\alpha_n}$ even does not satisfy the Robbins-Monro condition and we need the more recent ODE approach from \citet{lauand2024revisiting}.

\section{Conclusion}
This paper provides the first formalization of the  almost sure convergence of linear TD and $Q$-learning,
significantly advancing the state of the art in formalizing RL theory.
The developed framework is immediately ready to formalize more convergent RL results.
By taking conditional expectations on both sides of~\eqref{eq fundamental ineq} and telescoping, 
we can immediately get convergence rates in $\mathcal{L}_2$ with i.i.d. samples.
To get $\mathcal{L}_2$ convergence rates with Markovian samples, one can apply the technique from~\citet{srikant2019finite}.
Following \citet{qian2024almost,liu2026almost},
we can also obtain almost sure convergence rates easily under Markovian samples,
after getting a nonasymptotic version of the Robbins-Siegmund theorem following \citet{liu2022almost,karandikar2024convergence}.
We believe the aforementioned formalization shall be straightforward.
Next is concentration with exponential tails and $\mathcal{L}_p$ convergence. For this, we will need techniques from \citet{chen2025concentration} for i.i.d. samples and \citet{qian2024almost} for Markovian samples.
Both should be straightforward if we can formalize Hoeffding's lemma.
Our framework can be extended to other off-policy TD methods as well.
The family of gradient TD methods \citep{sutton2008convergent,sutton2009fast,yu2017convergence,zhang2021average,qian2025revisiting} is straightforward to formalize,
suppose no eligibility trace is involved.
The family of emphatic TD methods \citep{yu2015convergence,sutton2016emphatic} is much harder unless the trace is truncated \citep{zhang2022truncated} as the full trace will inevitably incur analysis of very complicated general state space Markov chains.
More challenging are the algorithms involving time-inhomogeneous Markov chains, e.g., (linear) SARSA \citep{zou2019sarsa}, (linear) $Q$-learning with a changing behavior policy \citep{meyn2024projected,liu2025linearq,liu2025extensions}, and policy gradient methods \citep{sutton1999policy,konda2002thesis,agarwal2019optimality,mei2020global,zhang2022global}.
For those algorithms,
the Markov chain is deeply coupled with the iterates and we envision major updates of our framework are necessary.

The most technically challenging part of this project comes from conditional expectation.
For example, consider computing $\E\qty[G(w_{t_m}, Y_{t+1})|\fF_{t_m}]$ where $\qty{Y_t}$ is a finite Markov chain and we use a matrix $P_\fY$ to denote its transition kernel.
It is straightforward for humans to conclude that 
\begin{align}
  \label{eq cond exp}
  \textstyle \E\qty[G(w_{t_m}, Y_{t+1})|\fF_{t_m}] = \sum_{y} P_\fY^{t+1 - t_m}(Y_{t_m}, y) G(w_{t_m}, y).
\end{align}
But to formalize this in Lean is highly nontrivial.
The difficulties come from two aspects.
First,
the conditional expectation in Lean is defined in an abstract and measure-theoretic way.
So many intuitive results about conditional expectation are highly nontrivial to formalize.
Second,
the probability space used for this measure-theoretic definition of conditional expectation is generated by the Ionescu-Tulcea theorem,
which is formalized in Lean in a constructive way for a generic family of history-dependent kernels.
So here we have to go into the details of the construction and simplify it for a Markov chain.
Formalizing~\eqref{eq cond exp} takes roughly 1,000 lines of Lean code, about 10\% of the entire project.
As a reference, \citet{sonoda2025lean} explicitly state that they use a customized proof in FoML to entirely avoid conditional expectation.
Addressing such small last-mile friction in formalization is a promising direction for future work.
And recent works such as \citet{liu2026mathliblemma,xie2026mathlibpr} have made some initial progress in this direction.

\subsection*{Acknowledgements}
This work is supported in part by the US National Science Foundation under the awards III-2128019, SLES-2331904, and CAREER-2442098, the Commonwealth Cyber Initiative's Central Virginia Node under the award VV-1Q26-001, a Cisco Faculty Research Award, and 4-VA at UVA.

%% file: appendix.tex

\appendix

\section{Lean and Mathlib in a Nutshell}
\label{sec bkg lean}
Lean is an interactive theorem prover based on dependent type theory: every theorem is a type, and a proof is a term inhabiting that type \citep{moura2021lean}.
Lean's trusted kernel checks proofs by typechecking, so correctness reduces to a small, auditable code base.
In practice, users rarely write raw proof terms.
Instead, Lean provides 
a tactic framework that lets users build proofs step-by-step using high level commands (e.g., \texttt{simp}, \texttt{rw}, \texttt{apply}).
Mathlib is Lean's flagship community-maintained library of formalized mathematics \citep{mathlib2020}.
It supplies definitions, theorems, and a large ecosystem of automation and conventions.
For readers unfamiliar with proof assistants, the key takeaway is: Lean is reliable because the kernel checks every detail, but writing a formal proof in Lean that goes through depends heavily on handling all the details in the proof carefully, even if they may look trivial to humans.

It is useful to separate two workflows in Lean. 
\emph{(1) Formalizing existing statements:} starting from an informal theorem statement in a paper or textbook and producing a well-typed Lean statement (without proof) that captures the intended meaning using Mathlib's definitions and conventions.
This step requires many design choices (which definitions to use, which generality and typeclass assumptions are idiomatic, which normal forms are preferred).
In general, there is no way to automatically check whether the formalization is correct and we have to resort to human experts.
\emph{(2) Proving a Lean statement:} given a type-checked Lean statement, producing a Lean proof that the kernel accepts.
The correctness can be automatically checked by Lean compiler,
which is the major advantage of formalized mathematics.
This work touches (1) and (2) in Section~\ref{sec statement} and Section~\ref{sec proof}, respectively.

\section{Proof of~\eqref{eq q lyapunovCandidate}}
\label{sec proof of eq q lyapunovCandidate}
\begin{align}
  &\textstyle \indot{\nabla \frac{1}{2} \norm{q - q_*}_p^2}{f(q) - q} \\
  =& \textstyle \indot{\nabla \frac{1}{2} \norm{q - q_*}_p^2}{\fT_*' q - q_*} + \indot{\nabla \frac{1}{2} \norm{q - q_*}_p^2}{q_* - q} \\
  =& \textstyle \indot{\nabla \frac{1}{2} \norm{q - q_*}_p^2}{\fT_*' q - q_*} - \norm{q - q_*}^2_p \explain{By computation} \\
  \leq& \textstyle \norm{\nabla \frac{1}{2} \norm{q - q_*}_p^2}_{(1 - p^{-1})^{-1}}\norm{\fT_*' q - q_*}_p - \norm{q - q_*}^2_p  \explain{By Holder's inequality} \\
  =& \norm{q - q_*}_p\norm{\fT_*' q - q_*}_p - \norm{q - q_*}^2_p  \explain{By computation} \\
  \leq& (\ns\na)^{1/p} \norm{q - q_*}_p\norm{\fT_*' q - q_*}_\infty - \norm{q - q_*}^2_p \explain{By norm equivalence} \\
  \leq& \gamma'(\ns\na)^{1/p} \norm{q - q_*}_p\norm{q - q_*}_\infty - \norm{q - q_*}^2_p  \\
  \leq& -(1 - \gamma'(\ns\na)^{1/p}) \norm{q - q_*}^2_p. \explain{By norm equivalence}
\end{align}

%% file: rlc.bbl
\begin{thebibliography}{80}
\providecommand{\natexlab}[1]{#1}
\providecommand{\url}[1]{\texttt{#1}}
\expandafter\ifx\csname urlstyle\endcsname\relax
  \providecommand{\doi}[1]{DOI: #1}\else
  \providecommand{\doi}{DOI: \begingroup \urlstyle{rm}\Url}\fi

\bibitem[Agarwal et~al.(2020)Agarwal, Kakade, Lee, and Mahajan]{agarwal2019optimality}
Alekh Agarwal, Sham~M. Kakade, Jason~D. Lee, and Gaurav Mahajan.
\newblock Optimality and approximation with policy gradient methods in markov decision processes.
\newblock In \emph{Proceedings of the Conference on Learning Theory}, 2020.

\bibitem[Antrobius \& Zhang(2026)Antrobius and Zhang]{antrobius2026divergence}
David Antrobius and Shangtong Zhang.
\newblock On the divergence of differential temporal difference learning without local clocks.
\newblock \emph{arXiv Preprint}, 2026.

\bibitem[Azar et~al.(2011)Azar, Munos, Ghavamzadeh, and Kappen]{azar2011speedy}
Mohammad~Gheshlaghi Azar, Remi Munos, Mohammad Ghavamzadeh, and Hilbert Kappen.
\newblock Speedy {Q-learning}.
\newblock In \emph{Advances in Neural Information Processing Systems}, 2011.

\bibitem[Beck \& Srikant(2012)Beck and Srikant]{beck2012error}
Carolyn~L Beck and Rayadurgam Srikant.
\newblock Error bounds for constant step-size {Q-learning}.
\newblock \emph{Systems \& Control Letters}, 2012.

\bibitem[Bellman(1957)]{bellman1957markovian}
Richard Bellman.
\newblock A markovian decision process.
\newblock \emph{Journal of Mathematics and Mechanics}, 1957.

\bibitem[Benveniste et~al.(1990)Benveniste, M{\'{e}}tivier, and Priouret]{benveniste1990MP}
Albert Benveniste, Michel M{\'{e}}tivier, and Pierre Priouret.
\newblock \emph{Adaptive Algorithms and Stochastic Approximations}.
\newblock Springer, 1990.

\bibitem[Bertsekas \& Tsitsiklis(1996)Bertsekas and Tsitsiklis]{bertsekas1996neuro}
Dimitri~P Bertsekas and John~N Tsitsiklis.
\newblock \emph{Neuro-Dynamic Programming}.
\newblock Athena Scientific Belmont, MA, 1996.

\bibitem[Bhandari et~al.(2018)Bhandari, Russo, and Singal]{bhandari2018finite}
Jalaj Bhandari, Daniel Russo, and Raghav Singal.
\newblock A finite time analysis of temporal difference learning with linear function approximation.
\newblock In \emph{Proceedings of the Conference on Learning Theory}, 2018.

\bibitem[Blaser \& Zhang(2026)Blaser and Zhang]{blaser2026asymptotic}
Ethan Blaser and Shangtong Zhang.
\newblock Asymptotic and finite sample analysis of nonexpansive stochastic approximations with markovian noise.
\newblock In \emph{Proceedings of the {AAAI} Conference on Artificial Intelligence}, 2026.

\bibitem[Blaser et~al.(2026)Blaser, Wang, and Zhang]{blaser2026almost}
Ethan Blaser, Jiuqi Wang, and Shangtong Zhang.
\newblock Almost sure convergence of differential temporal difference learning for average reward markov decision processes.
\newblock In \emph{Proceedings of the International Conference on Artificial Intelligence and Statistics}, 2026.

\bibitem[Borkar et~al.(2025)Borkar, Chen, Devraj, Kontoyiannis, and Meyn]{borkar2025ode}
Vivek Borkar, Shuhang Chen, Adithya Devraj, Ioannis Kontoyiannis, and Sean Meyn.
\newblock The {ODE} method for asymptotic statistics in stochastic approximation and reinforcement learning.
\newblock \emph{The Annals of Applied Probability}, 2025.

\bibitem[Borkar(2009)]{borkar2009stochastic}
Vivek~S Borkar.
\newblock \emph{Stochastic Approximation: A Dynamical Systems Viewpoint}.
\newblock Hindustan Book Agency, 2009.

\bibitem[Chen et~al.(2024)Chen, Maguluri, Shakkottai, and Shanmugam]{chen2024lyapunov}
Zaiwei Chen, Siva~T Maguluri, Sanjay Shakkottai, and Karthikeyan Shanmugam.
\newblock A lyapunov theory for finite-sample guarantees of markovian stochastic approximation.
\newblock \emph{Operations Research}, 2024.

\bibitem[Chen et~al.(2025)Chen, Maguluri, and Zubeldia]{chen2025concentration}
Zaiwei Chen, Siva~Theja Maguluri, and Martin Zubeldia.
\newblock Concentration of contractive stochastic approximation: Additive and multiplicative noise.
\newblock \emph{The Annals of Applied Probability}, 2025.

\bibitem[Chevallier(2024)]{chevallier2024verification}
Mark Chevallier.
\newblock \emph{Verification using formalised mathematics and theorem proving of reinforcement and deep learning}.
\newblock PhD thesis, The University of Edinburgh, 2024.

\bibitem[Chevallier \& Fleuriot(2021)Chevallier and Fleuriot]{chevallier2021formalising}
Mark Chevallier and Jacques Fleuriot.
\newblock Formalising the foundations of discrete reinforcement learning in isabelle/{HOL}.
\newblock \emph{arXiv Preprint}, 2021.

\bibitem[Dayan(1992)]{dayan1992convergence}
Peter Dayan.
\newblock The convergence of {TD}($\lambda$) for general $\lambda$.
\newblock \emph{Machine Learning}, 1992.

\bibitem[de~Moura \& Ullrich(2021)de~Moura and Ullrich]{moura2021lean}
Leonardo de~Moura and Sebastian Ullrich.
\newblock The {Lean} 4 theorem prover and programming language.
\newblock In \emph{International Conference on Automated Deduction}, 2021.

\bibitem[Degris et~al.(2012)Degris, White, and Sutton]{degris2012off}
Thomas Degris, Martha White, and Richard~S. Sutton.
\newblock Off-policy actor-critic.
\newblock In \emph{Proceedings of the International Conference on Machine Learning}, 2012.

\bibitem[Dvoretzky(1955)]{dvoretsky1955stochastic}
Aryeh Dvoretzky.
\newblock \emph{On stochastic approximation}.
\newblock Mathematics Division, Office of Scientific Research, US Air Force, 1955.

\bibitem[Even-Dar et~al.(2003)Even-Dar, Mansour, and Bartlett]{even2003learning}
Eyal Even-Dar, Yishay Mansour, and Peter Bartlett.
\newblock Learning rates for {Q-learning}.
\newblock \emph{Journal of Machine Learning Research}, 2003.

\bibitem[Jaakkola et~al.(1993)Jaakkola, Jordan, and Singh]{jaakkola1993convergence}
Tommi Jaakkola, Michael Jordan, and Satinder Singh.
\newblock Convergence of stochastic iterative dynamic programming algorithms.
\newblock In \emph{Advances in Neural Information Processing Systems}, 1993.

\bibitem[Karandikar \& Vidyasagar(2024)Karandikar and Vidyasagar]{karandikar2024convergence}
Rajeeva~Laxman Karandikar and Mathukumalli Vidyasagar.
\newblock Convergence rates for stochastic approximation: Biased noise with unbounded variance, and applications.
\newblock \emph{Journal of Optimization Theory and Applications}, 2024.

\bibitem[Kearns \& Singh(1998)Kearns and Singh]{kearns1998finite}
Michael Kearns and Satinder Singh.
\newblock Finite-sample convergence rates for {Q}-learning and indirect algorithms.
\newblock In \emph{Advances in Neural Information Processing Systems}, 1998.

\bibitem[Konda(2002)]{konda2002thesis}
Vijay~R. Konda.
\newblock \emph{Actor-Critic Algorithms}.
\newblock PhD thesis, Massachusetts Institute of Technology, 2002.

\bibitem[Kushner \& Yin(2003)Kushner and Yin]{kushner2003stochastic}
Harold Kushner and G~George Yin.
\newblock \emph{Stochastic approximation and recursive algorithms and applications}.
\newblock Springer, 2003.

\bibitem[Lakshminarayanan \& Szepesv{\'{a}}ri(2018)Lakshminarayanan and Szepesv{\'{a}}ri]{lakshminarayanan2018linear}
Chandrashekar Lakshminarayanan and Csaba Szepesv{\'{a}}ri.
\newblock Linear stochastic approximation: How far does constant step-size and iterate averaging go?
\newblock In \emph{Proceedings of the International Conference on Artificial Intelligence and Statistics}, 2018.

\bibitem[Lauand \& Meyn(2024)Lauand and Meyn]{lauand2024revisiting}
Caio~Kalil Lauand and Sean Meyn.
\newblock Revisiting step-size assumptions in stochastic approximation.
\newblock \emph{arXiv Preprint}, 2024.

\bibitem[Lee \& He(2020)Lee and He]{lee2019unified}
Donghwan Lee and Niao He.
\newblock A unified switching system perspective and convergence analysis of {Q-Learning} algorithms.
\newblock In \emph{Advances in Neural Information Processing Systems}, 2020.

\bibitem[Li et~al.(2024{\natexlab{a}})Li, Wang, He, Wu, Xu, and Wen]{li2024formalization}
Chenyi Li, Ziyu Wang, Wanyi He, Yuxuan Wu, Shengyang Xu, and Zaiwen Wen.
\newblock Formalization of complexity analysis of the first-order optimization algorithms.
\newblock \emph{arXiv Preprint}, 2024{\natexlab{a}}.

\bibitem[Li et~al.(2025{\natexlab{a}})Li, Wang, Bai, Duan, Gao, Hao, and Wen]{li2025formalizationa}
Chenyi Li, Zichen Wang, Yifan Bai, Yunxi Duan, Yuqing Gao, Pengfei Hao, and Zaiwen Wen.
\newblock Formalization of algorithms for optimization with block structures.
\newblock \emph{arXiv Preprint}, 2025{\natexlab{a}}.

\bibitem[Li et~al.(2025{\natexlab{b}})Li, Xu, Sun, Zhou, and Wen]{li2025formalizationb}
Chenyi Li, Shengyang Xu, Chumin Sun, Li~Zhou, and Zaiwen Wen.
\newblock Formalization of optimality conditions for smooth constrained optimization problems.
\newblock \emph{arXiv Preprint}, 2025{\natexlab{b}}.

\bibitem[Li et~al.(2020)Li, Wei, Chi, Gu, and Chen]{li2020sample}
Gen Li, Yuting Wei, Yuejie Chi, Yuantao Gu, and Yuxin Chen.
\newblock Sample complexity of asynchronous {Q}-learning: Sharper analysis and variance reduction.
\newblock In \emph{Advances in Neural Information Processing Systems}, 2020.

\bibitem[Li et~al.(2024{\natexlab{b}})Li, Cai, Chen, Wei, and Chi]{li2024q}
Gen Li, Changxiao Cai, Yuxin Chen, Yuting Wei, and Yuejie Chi.
\newblock Is {Q-learning} minimax optimal? a tight sample complexity analysis.
\newblock \emph{Operations Research}, 2024{\natexlab{b}}.

\bibitem[Liu \& Yuan(2022)Liu and Yuan]{liu2022almost}
Jun Liu and Ye~Yuan.
\newblock On almost sure convergence rates of stochastic gradient methods.
\newblock In \emph{Proceedings of the Conference on Learning Theory}, 2022.

\bibitem[Liu et~al.(2025{\natexlab{a}})Liu, Chen, and Zhang]{liu2025ode}
Shuze Liu, Shuhang Chen, and Shangtong Zhang.
\newblock The {ODE} method for stochastic approximation and reinforcement learning with markovian noise.
\newblock \emph{Journal of Machine Learning Research}, 2025{\natexlab{a}}.

\bibitem[Liu et~al.(2025{\natexlab{b}})Liu, Xie, and Zhang]{liu2025extensions}
Xinyu Liu, Zixuan Xie, and Shangtong Zhang.
\newblock Extensions of {Robbins-Siegmund} theorem with applications in reinforcement learning.
\newblock \emph{arXiv Preprint}, 2025{\natexlab{b}}.

\bibitem[Liu et~al.(2025{\natexlab{c}})Liu, Xie, and Zhang]{liu2025linearq}
Xinyu Liu, Zixuan Xie, and Shangtong Zhang.
\newblock Linear ${Q}$-learning does not diverge in ${L}^2$: Convergence rates to a bounded set.
\newblock In \emph{Proceedings of the International Conference on Machine Learning}, 2025{\natexlab{c}}.

\bibitem[Liu et~al.(2026{\natexlab{a}})Liu, Xie, Moeini, Chen, Liu, Meng, Zhang, and Zhang]{liu2026mathliblemma}
Xinyu Liu, Zixuan Xie, Amir Moeini, Claire Chen, Shuze Liu, Yu~Meng, Aidong Zhang, and Shangtong Zhang.
\newblock Mathliblemma: Folklore lemma generation and benchmark for formal mathematics.
\newblock In \emph{Proceedings of the International Conference on Machine Learning}, 2026{\natexlab{a}}.

\bibitem[Liu et~al.(2026{\natexlab{b}})Liu, Xie, and Zhang]{liu2026almost}
Xinyu Liu, Zixuan Xie, and Shangtong Zhang.
\newblock Almost sure convergence rates of stochastic approximation and reinforcement learning via a {Poisson-Moreau} drift.
\newblock \emph{arXiv Preprint}, 2026{\natexlab{b}}.

\bibitem[Marion(2025)]{marion2025formalization}
Etienne Marion.
\newblock A formalization of the ionescu-tulcea theorem in mathlib.
\newblock \emph{arXiv Preprint}, 2025.

\bibitem[Math-Inc(2025)]{mathinc2025}
Math-Inc.
\newblock Introducing {Gauss}, an agent for autoformalization.
\newblock \emph{\url{https://www.math.inc/gauss}}, 2025.

\bibitem[Mei et~al.(2020)Mei, Xiao, Szepesv{\'{a}}ri, and Schuurmans]{mei2020global}
Jincheng Mei, Chenjun Xiao, Csaba Szepesv{\'{a}}ri, and Dale Schuurmans.
\newblock On the global convergence rates of softmax policy gradient methods.
\newblock In \emph{Proceedings of the International Conference on Machine Learning}, 2020.

\bibitem[Meyn(2024)]{meyn2024projected}
Sean Meyn.
\newblock The projected bellman equation in reinforcement learning.
\newblock \emph{IEEE Transactions on Automatic Control}, 2024.

\bibitem[Puterman(1994)]{puterman2014markov}
Martin~L Puterman.
\newblock \emph{{Markov} decision processes: discrete stochastic dynamic programming}.
\newblock John Wiley \& Sons, 1994.

\bibitem[Qian \& Zhang(2025)Qian and Zhang]{qian2025revisiting}
Xiaochi Qian and Shangtong Zhang.
\newblock Revisiting a design choice in gradient temporal difference learning.
\newblock In \emph{Proceedings of the International Conference on Learning Representations}, 2025.

\bibitem[Qian et~al.(2024)Qian, Xie, Liu, and Zhang]{qian2024almost}
Xiaochi Qian, Zixuan Xie, Xinyu Liu, and Shangtong Zhang.
\newblock Almost sure convergence rates and concentration of stochastic approximation and reinforcement learning with markovian noise.
\newblock \emph{arXiv Preprint}, 2024.

\bibitem[Qu \& Wierman(2020)Qu and Wierman]{qu2020finite}
Guannan Qu and Adam Wierman.
\newblock Finite-time analysis of asynchronous stochastic approximation and $ q $-learning.
\newblock In \emph{Proceedings of the Conference on Learning Theory}, 2020.

\bibitem[Robbins \& Siegmund(1971)Robbins and Siegmund]{robbins1971convergence}
Herbert Robbins and David Siegmund.
\newblock A convergence theorem for non negative almost supermartingales and some applications.
\newblock \emph{Optimizing Methods in Statistics}, 1971.

\bibitem[Sch{\"a}feller \& Abdulaziz(2022)Sch{\"a}feller and Abdulaziz]{schafeller2022formally}
Maximilian Sch{\"a}feller and Mohammad Abdulaziz.
\newblock Formally verified solution methods for infinite-horizon markov decision processes.
\newblock \emph{arXiv Preprint}, 2022.

\bibitem[Sch{\"a}ffeler \& Abdulaziz(2025)Sch{\"a}ffeler and Abdulaziz]{schaffeler2025formally}
Maximilian Sch{\"a}ffeler and Mohammad Abdulaziz.
\newblock Formally verified approximate policy iteration.
\newblock In \emph{Proceedings of the AAAI Conference on Artificial Intelligence}, 2025.

\bibitem[Shah \& Xie(2018)Shah and Xie]{shah2018q}
Devavrat Shah and Qiaomin Xie.
\newblock {Q}-learning with nearest neighbors.
\newblock In \emph{Advances in Neural Information Processing Systems}, 2018.

\bibitem[Sonoda et~al.(2025)Sonoda, Kasaura, Mizuno, Tsukamoto, and Onda]{sonoda2025lean}
Sho Sonoda, Kazumi Kasaura, Yuma Mizuno, Kei Tsukamoto, and Naoto Onda.
\newblock Lean formalization of generalization error bound by rademacher complexity.
\newblock \emph{arXiv Preprint}, 2025.

\bibitem[Srikant \& Ying(2019)Srikant and Ying]{srikant2019finite}
Rayadurgam Srikant and Lei Ying.
\newblock Finite-time error bounds for linear stochastic approximation and td learning.
\newblock In \emph{Proceedings of the Conference on Learning Theory}, 2019.

\bibitem[Sutton(1988)]{sutton1988learning}
Richard~S. Sutton.
\newblock Learning to predict by the methods of temporal differences.
\newblock \emph{Machine Learning}, 1988.

\bibitem[Sutton \& Barto(2018)Sutton and Barto]{sutton2018reinforcement}
Richard~S Sutton and Andrew~G Barto.
\newblock \emph{Reinforcement Learning: An Introduction (2nd Edition)}.
\newblock MIT press, 2018.

\bibitem[Sutton et~al.(1999)Sutton, McAllester, Singh, and Mansour]{sutton1999policy}
Richard~S Sutton, David McAllester, Satinder Singh, and Yishay Mansour.
\newblock Policy gradient methods for reinforcement learning with function approximation.
\newblock In \emph{Advances in Neural Information Processing Systems}, 1999.

\bibitem[Sutton et~al.(2008)Sutton, Szepesv{\'a}ri, and Maei]{sutton2008convergent}
Richard~S Sutton, Csaba Szepesv{\'a}ri, and Hamid Maei.
\newblock A convergent {$O (n)$} algorithm for off-policy temporal-difference learning with linear function approximation.
\newblock In \emph{Advances in Neural Information Processing Systems}, 2008.

\bibitem[Sutton et~al.(2009)Sutton, Maei, Precup, Bhatnagar, Silver, Szepesv{\'{a}}ri, and Wiewiora]{sutton2009fast}
Richard~S. Sutton, Hamid~Reza Maei, Doina Precup, Shalabh Bhatnagar, David Silver, Csaba Szepesv{\'{a}}ri, and Eric Wiewiora.
\newblock Fast gradient-descent methods for temporal-difference learning with linear function approximation.
\newblock In \emph{Proceedings of the International Conference on Machine Learning}, 2009.

\bibitem[Sutton et~al.(2016)Sutton, Mahmood, and White]{sutton2016emphatic}
Richard~S. Sutton, Ashique~Rupam Mahmood, and Martha White.
\newblock An emphatic approach to the problem of off-policy temporal-difference learning.
\newblock \emph{Journal of Machine Learning Research}, 2016.

\bibitem[{The mathlib Community}(2020)]{mathlib2020}
{The mathlib Community}.
\newblock The {Lean} mathematical library.
\newblock In \emph{Proceedings of the {ACM} {SIGPLAN} International Conference on Certified Programs and Proofs}, 2020.

\bibitem[Tsitsiklis(1994)]{tsitsiklis1994asynchronous}
John~N Tsitsiklis.
\newblock Asynchronous stochastic approximation and {Q-learning}.
\newblock \emph{Machine Learning}, 1994.

\bibitem[Tsitsiklis \& Roy(1999)Tsitsiklis and Roy]{tsitsiklis1999average}
John~N. Tsitsiklis and Benjamin~Van Roy.
\newblock Average cost temporal-difference learning.
\newblock \emph{Automatica}, 1999.

\bibitem[Tsitsiklis \& Van~Roy(1997)Tsitsiklis and Van~Roy]{tsitsiklis1997analysis}
John~N Tsitsiklis and Benjamin Van~Roy.
\newblock An analysis of temporal-difference learning with function approximation.
\newblock \emph{{IEEE} Transactions on Automatic Control}, 1997.

\bibitem[Tulcea(1949)]{tulcea1949mesures}
CT~Ionescu Tulcea.
\newblock Mesures dans les espaces produits.
\newblock \emph{Atti Accad. Naz. Lincei Rend}, 1949.

\bibitem[Vajjha et~al.(2021)Vajjha, Shinnar, Trager, Pestun, and Fulton]{vajjha2021certrl}
Koundinya Vajjha, Avraham Shinnar, Barry Trager, Vasily Pestun, and Nathan Fulton.
\newblock Certrl: formalizing convergence proofs for value and policy iteration in coq.
\newblock In \emph{Proceedings of the ACM SIGPLAN International Conference on Certified Programs and Proofs}, 2021.

\bibitem[Vajjha et~al.(2022)Vajjha, Trager, Shinnar, and Pestun]{vajjha2022formalization}
Koundinya Vajjha, Barry Trager, Avraham Shinnar, and Vasily Pestun.
\newblock Formalization of a stochastic approximation theorem.
\newblock \emph{arXiv Preprint}, 2022.

\bibitem[Wan et~al.(2021)Wan, Naik, and Sutton]{wan2020learning}
Yi~Wan, Abhishek Naik, and Richard~S. Sutton.
\newblock Learning and planning in average-reward markov decision processes.
\newblock In \emph{Proceedings of the International Conference on Machine Learning}, 2021.

\bibitem[Wang \& Zhang(2026)Wang and Zhang]{wang2024almost}
Jiuqi Wang and Shangtong Zhang.
\newblock Almost sure convergence of linear temporal difference learning with arbitrary features.
\newblock \emph{Journal of Machine Learning Research}, 2026.

\bibitem[Watkins \& Dayan(1992)Watkins and Dayan]{watkins1992q}
Christopher~JCH Watkins and Peter Dayan.
\newblock {Q}-learning.
\newblock \emph{Machine Learning}, 1992.

\bibitem[Watkins(1989)]{watkins1989learning}
Christopher John Cornish~Hellaby Watkins.
\newblock \emph{Learning from delayed rewards}.
\newblock PhD thesis, King's College, Cambridge, 1989.

\bibitem[Xie et~al.(2025)Xie, Liu, Chandra, and Zhang]{xie2025finite}
Zixuan Xie, Xinyu Liu, Rohan Chandra, and Shangtong Zhang.
\newblock Finite sample analysis of linear temporal difference learning with arbitrary features.
\newblock In \emph{Advances in Neural Information Processing Systems}, 2025.

\bibitem[Xie et~al.(2026)Xie, Liu, and Zhang]{xie2026mathlibpr}
Zixuan Xie, Xinyu Liu, and Shangtong Zhang.
\newblock Mathlib{PR}: Pull request merge-readiness benchmark for formal mathematical libraries.
\newblock \emph{arXiv Preprint}, 2026.

\bibitem[Yang et~al.(2025)Yang, Zhang, Cao, Zhou, Li, Guo, Yao, Chen, Li, and Ma]{yang2025formalml}
Xiao-Wen Yang, Zihao Zhang, Jianuo Cao, Zhi Zhou, Zenan Li, Lan-Zhe Guo, Yuan Yao, Taolue Chen, Yu-Feng Li, and Xiaoxing Ma.
\newblock Formalml: A benchmark for evaluating formal subgoal completion in machine learning theory.
\newblock \emph{arXiv Preprint}, 2025.

\bibitem[Yu(2015)]{yu2015convergence}
Huizhen Yu.
\newblock On convergence of emphatic temporal-difference learning.
\newblock In \emph{Proceedings of the Conference on Learning Theory}, 2015.

\bibitem[Yu(2017)]{yu2017convergence}
Huizhen Yu.
\newblock On convergence of some gradient-based temporal-differences algorithms for off-policy learning.
\newblock \emph{arXiv Preprint}, 2017.

\bibitem[Zhang \& Whiteson(2022)Zhang and Whiteson]{zhang2022truncated}
Shangtong Zhang and Shimon Whiteson.
\newblock Truncated emphatic temporal difference methods for prediction and control.
\newblock \emph{Journal of Machine Learning Research}, 2022.

\bibitem[Zhang et~al.(2021)Zhang, Wan, Sutton, and Whiteson]{zhang2021average}
Shangtong Zhang, Yi~Wan, Richard~S. Sutton, and Shimon Whiteson.
\newblock Average-reward off-policy policy evaluation with function approximation.
\newblock In \emph{Proceedings of the International Conference on Machine Learning}, 2021.

\bibitem[Zhang et~al.(2022)Zhang, Tachet~des Combes, and Laroche]{zhang2022global}
Shangtong Zhang, Remi Tachet~des Combes, and Romain Laroche.
\newblock Global optimality and finite sample analysis of softmax off-policy actor critic under state distribution mismatch.
\newblock \emph{Journal of Machine Learning Research}, 2022.

\bibitem[Zou et~al.(2019)Zou, Xu, and Liang]{zou2019sarsa}
Shaofeng Zou, Tengyu Xu, and Yingbin Liang.
\newblock Finite-sample analysis for {SARSA} with linear function approximation.
\newblock In \emph{Advances in Neural Information Processing Systems}, 2019.

\end{thebibliography}
